\documentclass[runningheads]{llncs}

\usepackage{accv}

\usepackage{accvabbrv}

\usepackage{graphicx}
\usepackage{booktabs}
\usepackage{multirow}
\usepackage{algorithm}
\usepackage{algpseudocode}

\newcommand{\ours}{ExpHOI\xspace}

\newcommand{\matInterCtrlal}[1]{\mathcal{#1}}

\usepackage[accsupp]{axessibility}

\usepackage[breaklinks,colorlinks,citecolor=accvblue]{hyperref}

\begin{document}

\title{Synthesizing Expressive Human-Object Interactions via Test-time Latent Adaptation}

\titlerunning{Expressive Human-Object Interactions via Test-time Latent Adaptation}

\author{Zijing Wu \and Yiping Li \and ZhiPeng Yu \and Yuzhou Zheng \and
Ziyazhuo Wang \and Binghui Zuo \and Yangang Wang\thanks{Corresponding author:
Yangang Wang. E-mail: \email{yangangwang@seu.edu.cn}. All authors are affiliated
with the Key Laboratory of Measurement and Control of Complex Systems of
Engineering, Ministry of Education, Nanjing, China.}}

\authorrunning{Z. Wu et al.}

\institute{Southeast University}

\maketitle

\begin{abstract}
In reality, when interacting with an object whose physical properties are unknown, humans sometimes go through an iterative adaptation process: they probe, over-exert, stagger, and adjust, rather than achieving a successful interaction on the first attempt. We argue that this adaptation process itself is an expressive aspect of human-object interaction, yet it has been largely overlooked in human motion synthesis. Both kinematic data-driven methods and physics-based controllers tend to produce flawlessly executed motions: the former lack sufficient data for learning fine-grained adaptation behaviors, while the latter, optimized through reinforcement learning (RL), are trained toward robust, deterministic behaviors by conditioning on privileged object dynamics. To synthesize such expressive motions, we propose a test-time latent adaptation framework called ExpHOI (\textbf{Exp}ressive \textbf{H}uman-\textbf{O}bject \textbf{I}nteraction), which is built upon an RL-trained physics-based control policy. During policy training, we encode diverse object dynamics contexts into a structured latent space. At test time, we use Advantage-Weighted Regression (AWR) to perform iterative latent optimization for a given object with unknown physical properties. During this process, temporary mismatches between the latent code and the true object dynamics turn latent optimization into an expressive progression from tentative, error-prone trials to proficient interaction. Adaptation behaviors inevitably come with occasional failures, so a high-level planner is introduced to handle failed cases to maintain full, consistent animations. We demonstrate our framework's effectiveness on a large-object transportation task, where such adaptation is especially pronounced. To train our framework, we introduce a mass-diverse object dataset with rich physical-property annotations. Across carrying and push-pull, our method pairs over 90\% success with markedly higher posture diversity, and is rated the most lifelike in our user study on animation liveliness.

\keywords{Physics-based character animation \and Human--object interaction \and Online adaptation \and Reinforcement learning \and Whole-body control}
\end{abstract}

\begin{figure}[t]
\vspace{-8pt}
    \centering
    \includegraphics[width=0.8\textwidth]{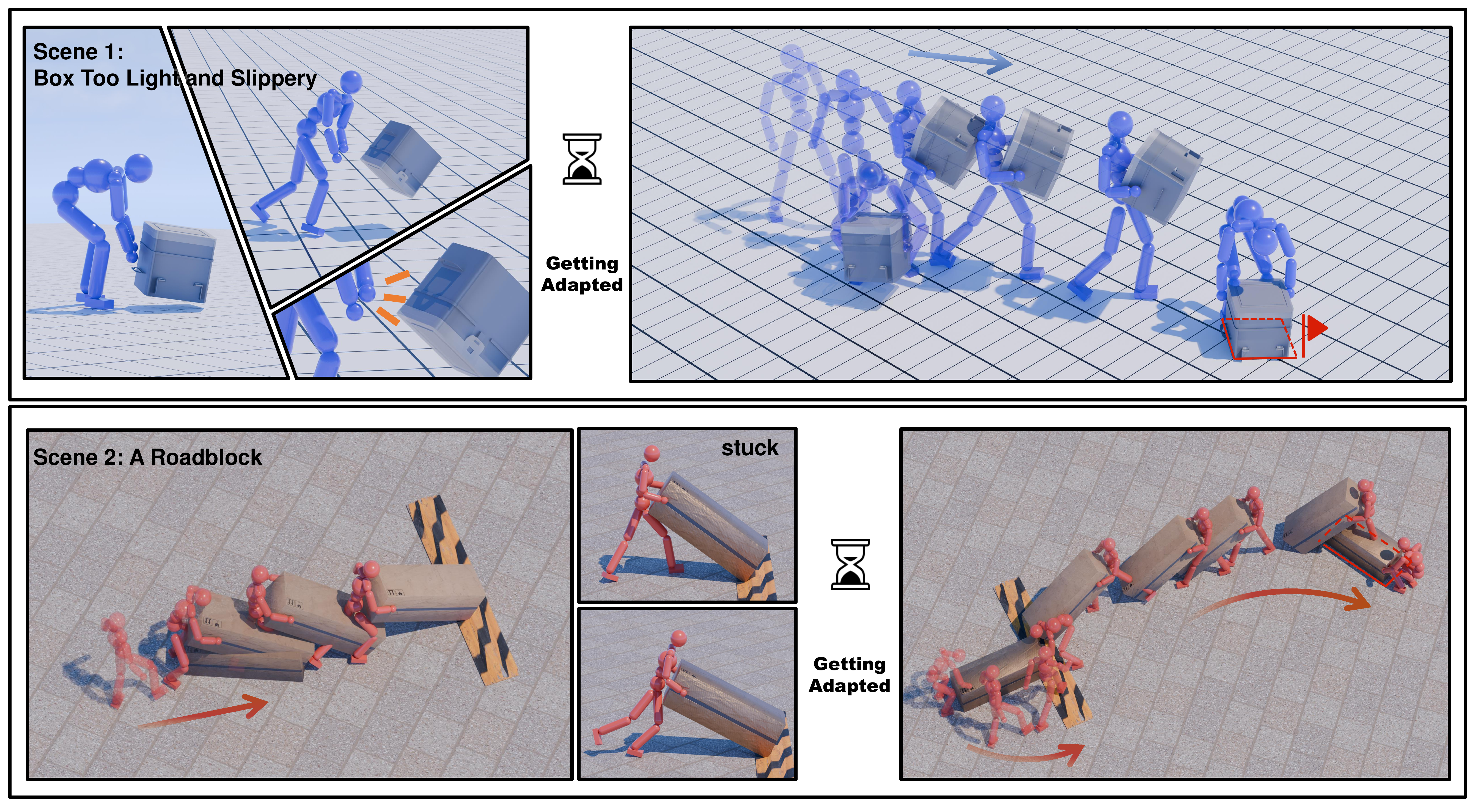}
    \caption{Illustration of \textbf{\ours}, a physics-based character animation framework, which emphasizes expressive animation during trial and error process.}
    \label{fig:teaser}
\end{figure}

\section{Introduction}
\label{sec:intro}

In reality, humans often misjudge an object's weight, mass distribution, or friction, so the interaction involves probing, over-exerting, staggering, sometimes failing, and then adjusting. This effortful adaptation is itself an expressive and valuable form of character animation, yet it has gone largely overlooked in computer animation. The probing and adaptation that arise under such physical uncertainty are among the scarcest and hardest-to-synthesize aspects of physical human--object interaction. We call this \textbf{expressive exertion}.

Existing methods struggle to produce such adaptive behavior, for three reasons. First, kinematic data-driven methods~\cite{dyn_caoHMP2020,dyn_TowardsDiverse,starke2019neural,dyn_li2023controllable,dyn_ron2025hoidinihumanobjectinteractiondiffusion} are inherently constrained by training data. Their generated motions are biased toward the distribution of training datasets~\cite{xu2023interdiff}. Motion capture of people straining under unknown or off-center loads is scarce, leading the learned models to produce only rigid, generic successful behaviors without expressive adaptive dynamics. Second, existing physics and RL-based controllers inherently generate optimal dynamics through reinforcement learning~\cite{box_gao2024coohoi,box_pan2025tokenhsi,InterPhysHassan2023,peng2022ase,tessler2023calm}. During interaction, the character is always equipped with full physical parameter knowledge, and tends to output the reward-optimal torques. This optimal control results in stiff motion and cannot simulate the hesitant, uncertain behaviors humans would show when underestimating or overestimating the object's mass or friction. Third, hand-crafting such adaptive interaction via keyframes is inefficient. For instance, animating a character that underestimates a box’s weight, overstrains, staggers, and readjusts its push demands extensive manual work. Meanwhile, this workflow easily compromises the physical consistency of generated movements.

To tackle these limitations, we take our cue from how people cope. Confronting an unfamiliar object, a person cannot know its weight, mass distribution, or friction in advance, and acts on a wrong estimate and unreliable perception of them. The natural human remedy is trial and error: attempt, sense the outcome, and refine the estimate over a few tries. We give the character the same strategy, keeping its physical estimate in a latent space and optimizing it over several trials with Advantage-Weighted Regression (AWR)~\cite{peng2019awr,RoboImitationPeng20}, which casts policy improvement as a supervised regression. The result is simple and intuitive: the character improves the way a person would, turning a familiar coping behavior into a controllable source of expressive motion.

To verify the effectiveness of our novel paradigm for generating expressive human-object animations, we evaluate our method on representative object transportation scenarios. We name our full system ExpHOI (\textbf{Exp}ressive \textbf{H}uman-\textbf{O}bject \textbf{I}nteraction), which couples a low-level object interaction controller with a high-level failure-aware planner. The object interaction controller is a physics-based policy conditioned on a latent code for physical properties. We train it on MDD (\textbf{M}assive-\textbf{D}iverse Object \textbf{D}ataset) with rich physical-property annotations. At test time, ground-truth physical properties of objects are unknown. The character initializes its latent embedding from a Gaussian distribution. Through a sequence of exploratory trials, our AWR scheme iteratively refines the latent estimation, during which, temporary discrepancies between latent codes and the real object dynamics drive latent optimization to produce an expressive behavioral progression: from hesitant, error-laden attempts to smooth, reliable object interaction. For this reason, we select representative intermediate latent outputs as the source of expressive motion animations and finalize the optimal latent parameters. Once the latent parameters are fixed, we integrate this low-level controller into our high-level task planner. Analogous to humans whose vivid behaviors stem from misestimation of object physical properties, expressive adaptive motions inevitably involve occasional failed attempts. Under such circumstances, the high-level planner is indispensable to activate a coherent give-up-and-retry behavioral cycle, ensuring physically complete and consistent animation sequences.

Our main contributions are as follows:
\begin{itemize}
    \renewcommand\labelitemi{\textbullet}
    \setlength{\itemsep}{4pt}
    \item \textbf{Latent test-time adaptation.} We propose a method that optimizes a latent estimate of the object's physical properties by advantage-weighted regression. The controllable intermediate trials are themselves expressive exertion.
    \item \textbf{The MDD dataset.} We build Mass-Diverse Dataset, a large-scale dataset of over $40{,}000$ procedurally generated objects with 13 physical-property labels, which fills the gap of physical-parameter annotation in existing datasets.
    \item \textbf{Failure-aware high-level planner.} We design a state-machine planner that coordinates the object interaction controller with a detouring controller, switching to a coherent give-up-adjust-retry cycle to ensure a complete animation system.

\end{itemize}

\section{Related Work}
\label{sec:related_work}

\noindent \textbf{Physics-based Human--Object Interaction.}
Many early studies on human--object interaction focused on kinematic generation, which often suffers from penetration, floating, and non-physical sliding, and overlooks physical plausibility~\cite{dyn_caoHMP2020,dyn_TowardsDiverse,dyn_li2023controllable,starke2019neural}. More recent work learns interaction control inside physics simulators~\cite{christen2022d,InterPhysHassan2023,l_wang2025sims}, combining reinforcement learning with imitation and motion priors to cover a wide range of plausible behaviors~\cite{box_gao2024coohoi,box_pan2025tokenhsi,peng2022ase,tessler2023calm}. Yet these methods mostly pursue interaction \emph{diversity} and overlook \emph{physical uncertainty}, even though object interaction is strongly shaped by complex, non-uniform physical properties. A few attempts at conveying effort and struggle rely on hand-crafted rules, force scripting, or musculoskeletal and fatigue models~\cite{komura2000musculoskeletal,fatigue_cheema2023discovering}, with limited expressiveness; effortful behavior under uncertain physics remains largely unexplored.

\noindent \textbf{Interaction with Dynamic Object.}
Many works target static scenes, such as sitting, lying down, or leaning against objects, and excel at generating natural contact and posture using scene-aware priors or diffusion-based synthesis~\cite{l_pan2024synthesizing,l_context_xiao2024unified}. These studies, however, largely ignore the feedback caused by changes in object state. Although research on dynamic objects is growing~\cite{basketball_wang2023physhoi,ttennis_su2025hitterhumanoidtabletennis,tennis_zhang2023vid2player3d,football_Liu2022MotorControl}, whole-body interaction with objects has mostly targeted small and lightweight items, such as a single chair, a rubber ball, or a tennis racket~\cite{xu2025intermimic,xu2025interact}. When extended to large or heavy objects, these approaches often cause instability or even falls.

\noindent \textbf{Contextual Interaction Adaptation.}
Perceiving and using context is crucial in human--object interaction. A common approach~\cite{zhang2024force,box_gao2024coohoi,box_pan2025tokenhsi} explicitly observes or encodes the context and feeds it to the policy as a conditional input. For example, FORCE~\cite{zhang2024force} puts parameters such as resistance and human force into an Intuitive Physics Encoding to refine action details; CooHOI~\cite{box_gao2024coohoi} adds object weight to the observation to increase load capacity. However, not all context is directly observable, which motivates a complementary line of implicit adaptation, typically through fine-tuning, meta-learning~\cite{MetaStrategy,pmlr-v70-finn17a}, or offline system identification~\cite{ramos2019bayessim,huang2023compass}. These methods still require network-parameter updates and often involve iterative training loops, which tend to be slow or twitchy in real-time animation. In contrast, we use smooth belief tuning based on AWR~\cite{RoboImitationPeng20}, which adapts without any network-parameter update or policy-gradient computation, and is better suited to generating continuous, natural adaptive animation.

\section{Methodology}
\label{sec:method}
\begin{figure*}[!t]
\vspace{-10pt}

  \hspace{-12pt}
   \includegraphics[width=1.05\linewidth]{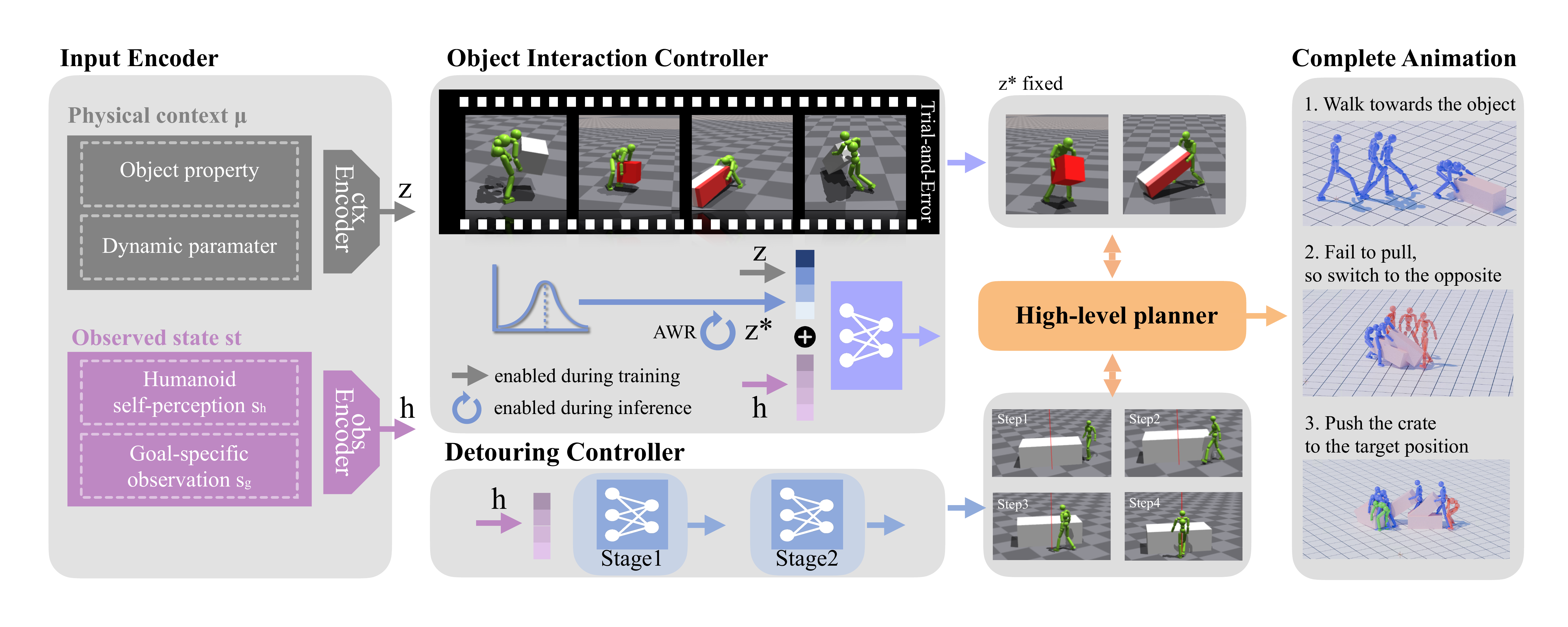}
   \caption{\textbf{Overview of \ours.} ExpHOI comprises two modules: Object Interaction Controller that carries and push--pulls the object, and Detouring Controller that handles obstacle avoidance. A high-level planner orchestrates both. The object interaction controller operates in two phases. During \emph{training}, a context encoder (ctxEncoder) produces a latent estimate $z$ of the object's physical properties. During \emph{inference}, the ctxEncoder is replaced by test-time latent tuning (AWR). With $z*$ fixed, the intermediate trials themselves yield the expressive exertion we extract.}\label{fig:pipeline}
\end{figure*}

To synthesize natural exertion, adaptation, and adjustment animation when a character faces objects of unknown physical properties, we propose ExpHOI, a physics-based animation framework with test-time latent adaptation. Sec.~\ref{symbols} models the physical properties that govern interaction and introduces our dataset. Sec.~\ref{InterCtrl} presents a base object interaction controller that endows the character with an implicit physical estimate in a cognitive latent space, so that it naturally exhibits reasonable exertion postures across diverse objects. Sec.~\ref{domainadaptation} introduces an information-bottleneck constraint at training time and an AWR estimate-tuning mechanism at inference time, which synthesizes the struggle-and-adaptation process and lets the character learn from failure. Sec.~\ref{planner} introduces a failure-aware high-level planner that tackles unexpected failure and produces coherent, multi-stage narrative animation.

\subsection{Problem Setup and the MDD Dataset} \label{symbols}
The real world contains numerous physical properties that influence how a character interacts with and exerts force on an object; they also determine the character's initial failures and subsequent improvements. To synthesize physically grounded animation, we first model the crucial physical information.

\noindent \textbf{Formulation.}
Let $\matInterCtrlal{M} = (\matInterCtrlal{S}, \matInterCtrlal{A}, \matInterCtrlal{T}, \matInterCtrlal{R}, \gamma)$ denote a Markov Decision Process, where $\matInterCtrlal{S}$ is the state space, $\matInterCtrlal{A}$ is the action space, $\matInterCtrlal{T}$ is the transition function, $\matInterCtrlal{R}$ is the reward function and $\gamma \in [0, 1)$ is the discount factor.
The state is formulated as: $s \triangleq (s_h, s_g, \mu)$, where $s_h$ represents the humanoid state, $s_g$ denotes the goal state, and $\mu$ stands for the crucial yet unknown context. The reward $r_t \in \mathbb{R}$ is composed of a style reward $r_t^{\text{style}}$ and a task reward $r_t^{\text{task}}$. The policy is designed to maximize the cumulative discounted reward $\sum_{t=0}^T \gamma^t r_t$. The Proximal Policy Optimization (PPO) algorithm \cite{schulman2017proximal} is employed to train the control policy model.

\noindent \textbf{State.} The character's self-perception $s_h$ consists of $\{s_h^{rh}, s_h^p, s_h^r, s_h^v, s_h^{av}\}$, where $s_h^{rh}$ denotes the global root height; $s_h^p$ represents the local body positions; $s_h^r$ stands for the relative rotations. $s_h^v$ is the local linear velocity; and $s_h^{av}$ is the local angular velocity. The object's state $s_o$ includes $\{s_o^p, s_o^r, s_o^v, s_o^{av}, s_o^{bbox}, s_o^{tar}\}$, where $s_o^{p}$, $s_o^{r}$, $s_o^{v}$, and $s_o^{av}$ respectively represent the position, rotation, linear velocity, and angular velocity of the object relative to the character's root node; $s_o^{bbox}$ denotes the object's bounding box in the local root coordinate frame; $s_o^{tar}$ denotes the target position and rotation. In the complete state $\{s_g, s_h, \mu\}$, \(s_g\) includes \(s_o\) and other task-specific states.

\noindent \textbf{Physical Context.} We focus on two types of physical context. The mathematical formulations are: (1) $\mu_o \in \mathbb{R}^{23}$, the physical properties of the transported object, including center-of-mass offset $\text{com}_{\text{offset}} \in \mathbb{R}^3$, inertia $\mathbf{I}_{\text{com}} \in \mathbb{R}^6$, principal moments of inertia $\mathbf{I}_{\text{principal}} \in \mathbb{R}^3$, quaternion for principal axis rotation $\mathbf{I}_{\text{rot}} \in \mathbb{R}^4$, total mass $m \in \mathbb{R}^1$, object contact and damping parameters $\boldsymbol{\theta}_{\text{obj}} \in \mathbb{R}^6$ (friction, rolling friction, torsion friction, restitution, angular damping, linear damping); (2) $\mu_e \in \mathbb{R}^4$, the dynamic parameters of the environmental interaction surface, corresponding to ground friction and restitution parameters $\boldsymbol{\theta}_{\text{ground}} \in \mathbb{R}^4$ that regulate contact behavior between the object, character and the ground.

\begin{figure}[t]
  \centering
  \begin{minipage}[t]{0.38\linewidth}
    \centering
    \includegraphics[width=\linewidth]{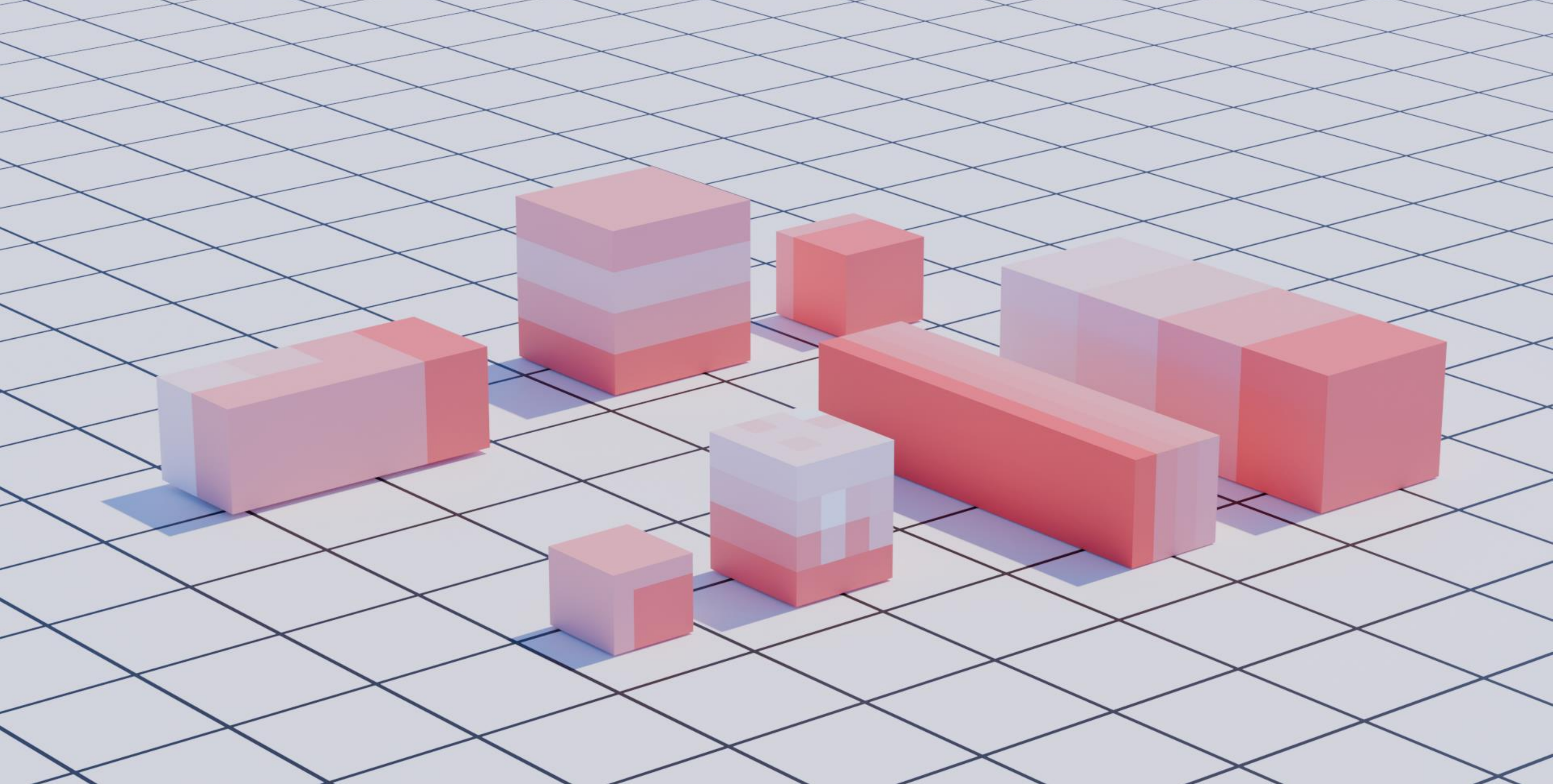}
  \end{minipage}
  \hspace{10pt}
  \begin{minipage}[t]{0.38\linewidth}
    \centering
    \includegraphics[width=\linewidth]{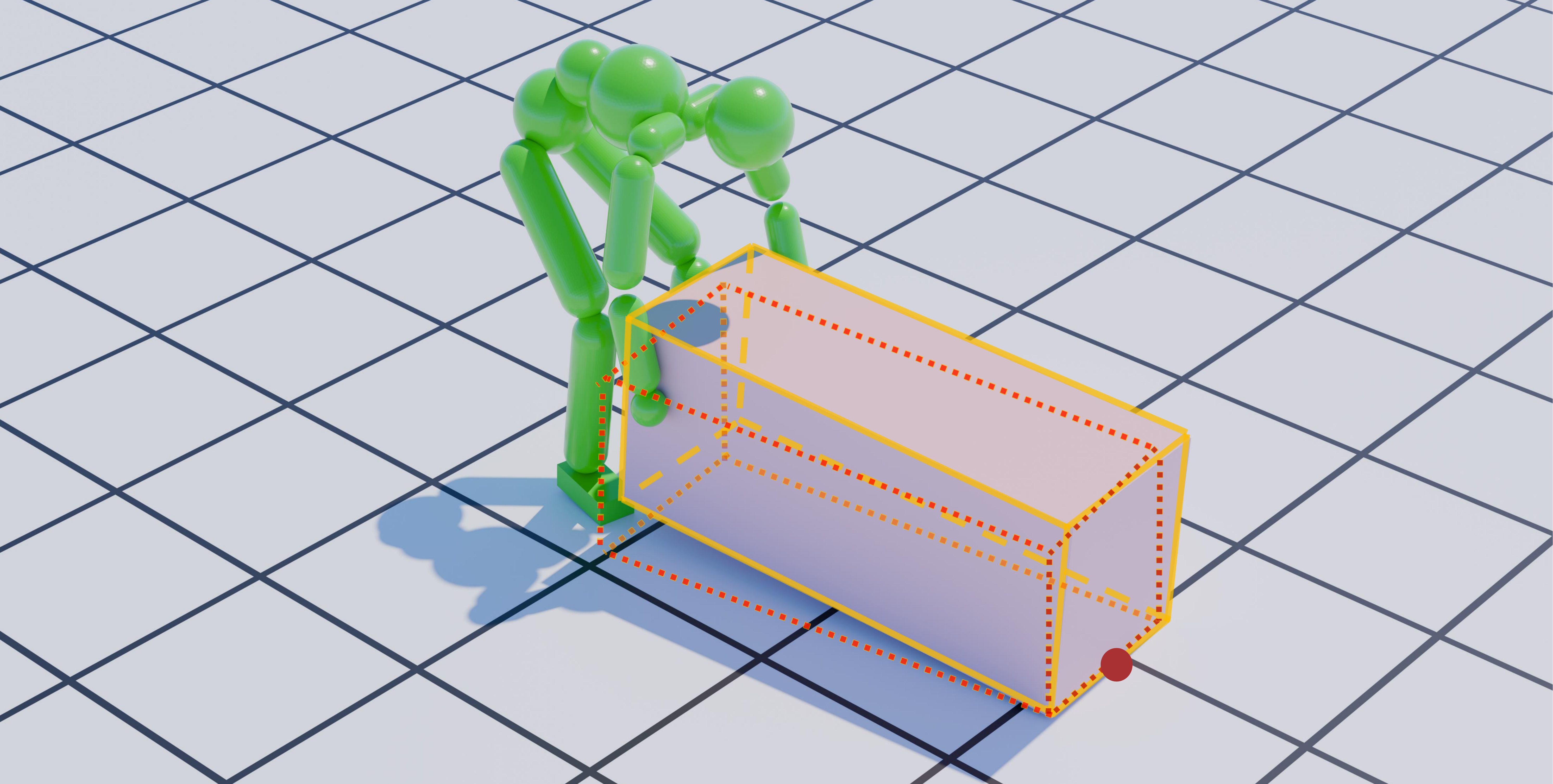}
  \end{minipage}
  \caption{MDD: lighter color means lighter weight (left). Ground point (right).}
  \label{fig:MDD_GP}
\end{figure}

\noindent \textbf{Mass-diverse Dataset.} Existing datasets typically lack such physical labels, so we propose a voxel-based mass-diverse dataset named MDD, which enables controllable mass properties, with each object annotated with 13 physical labels. Fig.~\ref{fig:MDD_GP} (left) shows typical objects, and lighter color means lighter weight. A detailed description is given in Sec.~\ref{data_init}.

\subsection{Object Interaction Controller} \label{InterCtrl}
In real interactions, humans hold an internal expectation about an object's friction, mass, and so on. To endow the virtual character with this cognitive ability, we compress the complex physical-world parameters into a continuous cognitive latent space, realized within the Object Interaction Controller (InterCtrl) that supports three skills: pushing, pulling, and carrying.

\noindent \textbf{Architecture.} The input to InterCtrl consists of the observable state \(s_t = \{s_{t}^h, s_{t}^{g}\}\) and the physical context \(\mu\), as illustrated in Fig.~\ref{fig:pipeline}. These are encoded by the observation encoder \(\epsilon_o\) and the context encoder \(\epsilon_c\) respectively, concatenated, and fed into an MLP that outputs joint target rotations to control the interaction. During training, \(\mu\) is encoded as a latent code \(z\) by \(\epsilon_c\), through which the character builds an estimate about the physical world and can handle environments with different physical properties. During inference, \(\epsilon_c\) is no longer used; instead, the optimal \(z^*\) is tuned from the latent space via a test-time adaptation method based on Advantage-Weighted Regression (AWR) and concatenated with the output of \(\epsilon_o\). This sense of exertion does not arise from feeding the character exact ground-truth parameters. Instead, it emerges during estimate updating and exploratory iteration. Throughout AWR optimization, discrepancies persist between intermediate latent embeddings and the true physical encodings of objects. Analogous to humans holding biased, inaccurate predictions of unknown objects at the initial trial stage, such imperfect latent estimations are the core source of expressive motion. Once we converge to a fixed latent code $z*$ that balances vivid adaptive movement and satisfactory task success rates, we freeze this latent variable and deploy the corresponding object interaction controller to produce full animation sequences. Since an imperfect $z*$ still carries a risk of failed manipulation, a high-level planner is introduced to coordinate the detouring controller and object interaction controller, resolving edge cases such as aborted attempts or objects jammed against obstacles. The combination of low-level tentative exploratory behaviors and high-level failure recovery mechanisms together yields authentic animations.

\noindent \textbf{Goal State and Reward.} State includes \(s_{t}^h\), \(s_{t}^g\) and \(\mu\), following the definition in Sec.~\ref{symbols}. In designing the goal state \(s_{t}^g\), we further introduce the held point and stand point inspired by CooHOI~\cite{box_gao2024coohoi}: the held point is the ideal midpoint between the two hands, and the stand point is the ideal standing position when hauling. The reward is formulated as \(r_t = w_G\,r^{\text{goal}}_t + w_S\,r^{\text{style}}_t\), where the style term \(r^{\text{style}}_t\) follows the Adversarial Motion Priors (AMP) paradigm~\cite{2021-TOG-AMP}. Aligning an object precisely is often challenging, especially for elongated, heavy crates whose position may be aligned while the bounding box remains misaligned. To solve this, we introduce the Ground Point (GP), defined as the midpoint of the lowest edge of the bounding box, as shown in Fig.~\ref{fig:MDD_GP}. The ground point reward is $r_{gp}=\exp\bigl(-\lambda \cdot \lVert GP_{xy} - GP_{xy}^{tar} \rVert^{2}\bigr)$, where $GP_{xy}$ is the current object ground point in the $XY$-plane, $GP_{xy}^{tar}$ is its target, and $\lambda$ controls sensitivity. With the GP aligned, the primary challenge is reduced from full-pose alignment to merely adjusting orientation, decoupling translational and rotational errors and easing final alignment for large and heavy objects.

\subsection{Synthesizing Expressive Exertion via Test-time Adaptation}\label{domainadaptation}
Section~\ref{InterCtrl} endowed the character with an estimate about the physical world and showed that exertion arises as it acts on that estimate while exploring. Making this practical requires shaping the estimate into a smooth, easily adjustable latent space at training time, and recovering a suitable estimate at inference time when the object's properties are unknown; we address these with an information-bottleneck (IB) constraint and AWR estimate tuning, respectively.

\noindent \textbf{Modeling Physical Context via Latent Space.} Recall from Sec.~\ref{InterCtrl} the context encoder $E(z|\mu) = \matInterCtrlal{N}(m(\mu), \text{diag}(s(\mu)^2))$ that maps the physical context $\mu$ into the estimate $z$ at training time. We follow~\cite{RoboImitationPeng20} and add an Information Bottleneck (IB) regularizer to the policy loss:
\begin{equation} \label{eq:ib_regularizer}
\text{IB} = \beta \cdot \text{KL}(q(z|\mu) \| \matInterCtrlal{N}(0, I)),
\end{equation}
where the KL divergence is
\begin{equation} \label{eq:ib_kl}
\text{KL} = 0.5 \sum \left( m^2 + s^2 - 2 \log s - 1 \right).
\end{equation}
This constrains the estimate encoding toward a simple normal distribution, compelling the policy to learn invariant, task-agnostic representations of physical essence rather than over-relying on spurious correlations with specific parameters \(\mu\). This not only enhances the policy’s robustness and generalization to unseen physical contexts but also establishes a foundation for the adaptive mechanism detailed in the following part.

\noindent \textbf{Tuning the Estimate via Trial-and-Error.} During inference, $\mu$ is unknown and $E(z|\mu)$ is unusable, so we let the character navigate this space through trials to find the optimal estimate $z^*$. Specifically, we use AWR~\cite{peng2019awr} to search for a $z^*$ that maximizes the expected cumulative discounted reward:
\begin{equation} \label{eq:optimal_z}
z^* = \arg\max_{z} \mathbb{E}_{\tau \sim p_{\text{new}}(\tau \mid \pi(s, z))} \left[ \sum_{t=0}^{T-1} \gamma^t r_t \right].
\end{equation}
Here \(p_{\text{new}}(\tau\mid\pi(s, z))\) denotes the trajectory distribution under uncertain physical context, \(\pi(\cdot \mid s, z)\) is the object interaction controller policy, $r_t$ is the reward at step $t$, \(\gamma\) is the discount factor, and $T$ is the episode length. Algorithm~\ref{alg:awr_adapt} illustrates the scheme. We initialize a prior \(\matInterCtrlal{D}_0(z) = \matInterCtrlal{N}(0, I)\) and enter a trial-and-error loop. For each trial $k$ we run \(\pi(s, z_k)\) over $B$ parallel simulation environments and record the batch-averaged return \(\bar R_k\). The sample pair \((z_k, \bar R_k)\) is then added to dataset $D$, and we calculate normalized weights \(w_i\), where samples yielding better return are assigned larger weights. We clamp the weights via \(w_i = \min(w_i, 3.0)\) to prevent extreme weights from dominating the update and to stabilize the latent distribution. After \(K\) trials, we take the mean of the converged latent distribution as $z^*$ and freeze this latent code for all subsequent inference. At this stage, the latent estimate may still retain mild misalignment with the ground-truth physical encoding, which balances high task success rates and natural, expressive exertion motions simultaneously. This is exactly the desired property of our pipeline. Humans exhibit vivid corrective adjustment behaviors precisely because they underestimate or overestimate physical properties such as mass and friction. Our model directly benefits from this controllable mismatch between latent estimations and ground-truth physical encodings introduced throughout the AWR optimization process.

It should be noted that in transportation, the instantaneous reward $r_t$ is strongly coupled with task progress and wall-clock time: a rollout that has already lifted and carried the box naturally receives a higher return than one still approaching it at the same timestamp. To decouple learning signals from progress confounds and expose the contribution of $z$ to physical adaptation, we adopt a progress-alignment strategy: we first fix the initial pose and rotation of the character and object during trial-driven updates, ensuring comparable task phases across episodes; after $z$ converges, we re-enable initial-pose and path randomization to assess and improve generalization under realistic execution variability.

\begin{algorithm}
\caption{AWR Adaptation Process}
\label{alg:awr_adapt}
\begin{algorithmic}[1]
\Require Trained policy $\pi(s, z)$, new physical context, temperature $\alpha$, max trials $K_{\max}$, max iteration $S$, batch size $B$
\State Initialize $\matInterCtrlal{D}_0(z) = \matInterCtrlal{N}(0, I)$, $D = \emptyset$
\For{$k = 0$ \textbf{to} $K_{\max} - 1$}
    \State Sample $z_k \sim \matInterCtrlal{D}_k(z)$
    \State Run episodes with $\pi(s, z_k)$ under new physical context, get batch-averaged return $\bar R_k = \frac{1}{B}\sum_{b=1}^B \sum_{t=0}^{T-1} \gamma^t r_{t,b}$

    \State Add $(z_k, \bar R_k)$ to $D$
    \State Compute $w_i = clamp\left(\exp\left( \frac{norm(R_i)}{\alpha} \right)\right)$
     \State Let $m_k, \Sigma_k$ denote mean and covariance of $\matInterCtrlal{D}_k(z)$
    \For{$s = 0$ \textbf{to} $S - 1$}
        \State $L = -\frac{1}{|D|} \sum_i w_i \log \matInterCtrlal{N}(z_i; m_k, \Sigma_k)$
        \State Update $m_k, \Sigma_k$ via SGD on $L$
    \EndFor
    \State $\matInterCtrlal{D}_{k+1}(z) = \matInterCtrlal{N}(m_{\text{new}}, \Sigma_{\text{new}})$
\EndFor
\State $z^* = m_{K}$
\end{algorithmic}
\end{algorithm}

\subsection{Failure-Aware High-level Planner} \label{planner}
Sections~\ref{InterCtrl} and~\ref{domainadaptation} let the character adapt its exertion under continuous physical uncertainty. Our core goal is to reproduce human-like behaviors of staggering, adaptation and correction stemming from inaccurate estimations of object properties, which inevitably leads to occasional failed attempts. Furthermore, some setbacks cannot be resolved by adjusting the estimate, for example being jammed by an obstacle, or having lifted one side of an object that still cannot be pushed or pulled. To keep the animation coherent and complete in these cases, we add a failure-aware high-level planner that coordinates the object interaction controller with a Detouring Controller (DtrCtrl), forming a full animation system capable of generating complete behavioral arcs featuring natural physical exertion and post-failure recovery adjustments.

\begin{figure}
  \centering
  \begin{subfigure}[t]{0.45\textwidth}
    \centering
    \includegraphics[width=\linewidth]{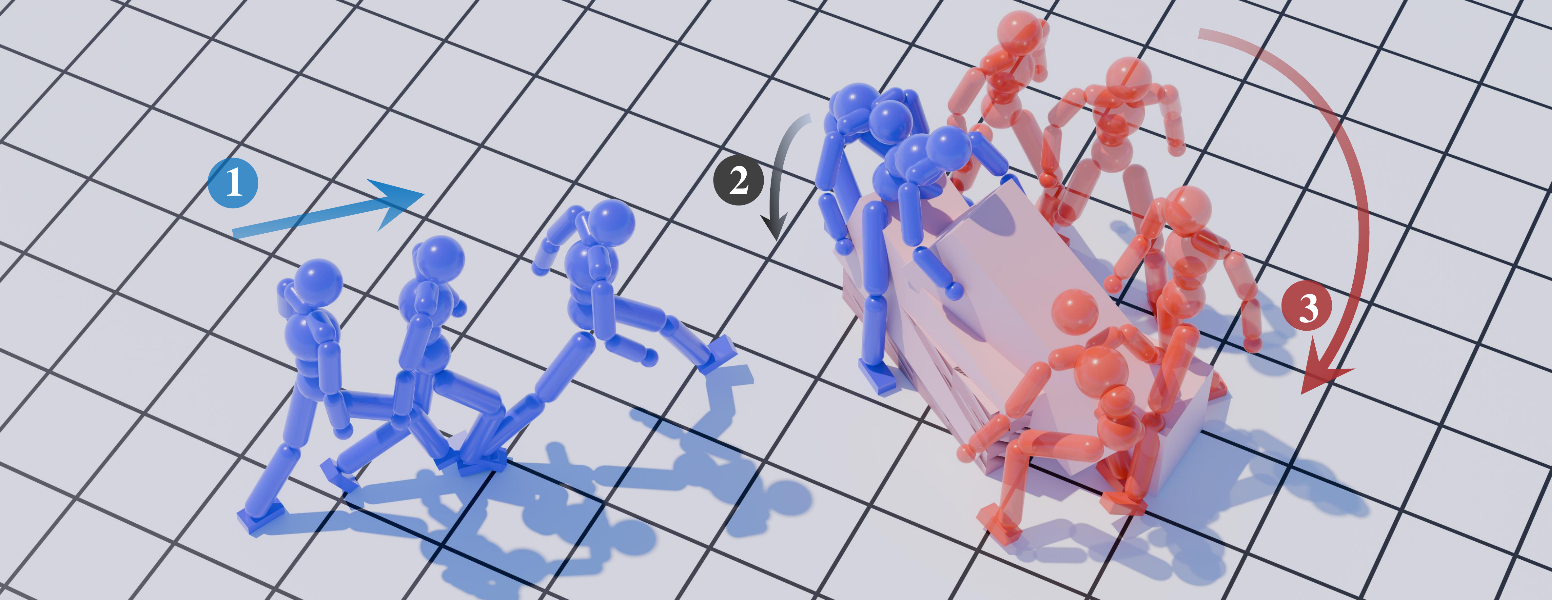}
    \caption{Pulling attempt fails, switching to the opposite side.}
    \label{fig:hlp_detour}
  \end{subfigure}
  \hfill
  \begin{subfigure}[t]{0.45\textwidth}
    \centering
    \includegraphics[width=\linewidth]{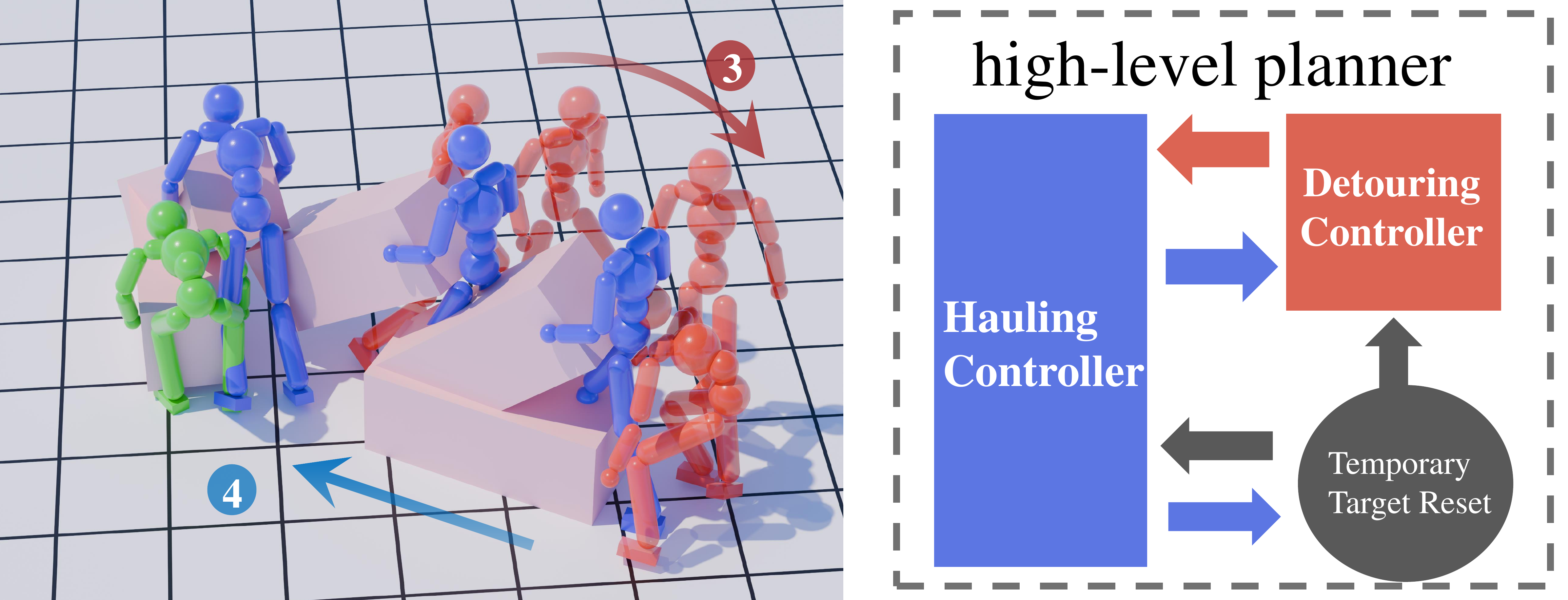}
    \caption{Push the object to the target position.}
    \label{fig:hlp_push}
  \end{subfigure}
  \caption{High-level planner mechanism.}
  \label{fig:high_level_planner}
\end{figure}

\noindent \textbf{Detouring Controller.} DtrCtrl performs short-range side-switching and detouring around individual obstacles, as indicated by Step 3 in Fig.~\ref{fig:hlp_detour}. Its observation is \( s_t \triangleq (s^h_t, s^g_t) \). Following NavRL~\cite{NavRL}, the goal state incorporates the target standing position and local Signed Distance Function (SDF) features to penalize collisions and undesired contacts. Because humanoid gait involves arm swinging and long strides, we include a core body--obstacle minimum-distance term to penalize potential brush contacts between limb endpoints and obstacle edges. The training of DtrCtrl faces significant exploration challenges, so we adopt curriculum learning: the first phase trains on small obstacles with standing positions restricted to the midpoints of the four sides, letting the policy master basic detouring and safe-distance control; the second phase expands to full-sized obstacles with standing positions placed arbitrarily around the obstacle, developing robust locomotion.

\noindent \textbf{Rule-based State Machine.} We propose a rule-based high-level planner to switch between InterCtrl and DtrCtrl, serving the scenario of side-switching and retry after hauling failures. Transitions are triggered by guard conditions: (1) a switch from InterCtrl to DtrCtrl occurs when a failure event is detected; (2) a switch back to InterCtrl is initiated upon reaching the target standing position. Fig.~\ref{fig:high_level_planner} shows the state transitions.

\noindent \textbf{Failure Detection and Temporary Target.} The failure-transition condition comprises an instantaneous failure check and a historical sliding window. The instantaneous check covers two cases: the humanoid is far from the target while both hands contact the object but the object stays stationary for an extended period; or the humanoid is far from the target and the object drops back to the ground after being lifted. A historical sliding window then filters false triggers: a failure is declared if the instantaneous check is satisfied multiple times within the window and no prior failure has been recorded. This debounced criterion mitigates chattering and duplicate triggers while remaining simple and largely insensitive to thresholds. For the special case where one side of the object is lifted but it cannot be pushed or pulled, direct switching would fail, since the humanoid remains in an intermediate contact state. To address this, we issue a temporary target that redirects the hauling target to the ground projection of the object's current position, driving InterCtrl to first place the object safely on the ground; after confirming the landing, the planner switches to DtrCtrl for safe detouring and side-switching.

\section{Experiments}\label{sec:experiment}
Our experiments answer two questions: \textbf{(Q1)} does belief tuning synthesize expressive, property-dependent exertion that prior adaptation schemes and our own non-adaptive baseline do not? \textbf{(Q2)} is the adaptation quantitatively effective, as measured by ablations and success-rate comparisons? We address \textbf{Q1} primarily through extensive qualitative results (Sec.~\ref{exp:q1}), which directly show the realism, the sense of effort, and the per-object posture differences of the generated motions, as well as their advantage over prior methods; we address \textbf{Q2} through quantitative metrics (Sec.~\ref{exp:q2}).

\subsection{Setup}\label{data_init}
\noindent\textbf{Dataset with Physical Labels.} The core challenge here is physical uncertainty, yet existing datasets typically lack physical labels. Moreover, most real-world datasets are created via 3D scanning, and the resulting non-watertight meshes make it difficult to reliably control mass properties in simulators such as Isaac Gym~\cite{sim_NEURIPSDATASETSANDBENCHMARKS2021_28dd2c79}. We therefore propose MDD, a large-scale mass-diverse dataset. We model objects as bounding-box hexahedrons on a $4\times4\times4$ voxel grid (Fig.~\ref{fig:MDD_GP}); by assigning varied masses to the 64 cells (empty regions approximated by a mass of $10^{-8}$), we programmatically generate URDF/JSONL files, each annotated with 13 physical labels (7 for size, mass, and inertia, plus surface friction, rolling friction, torsional friction, restitution, angular damping, and linear damping). We model three mass-distribution types: (1) uniform; (2) density increasing/decreasing along the $x$, $y$, or $z$ axis; (3) uniform base with randomly distributed upper mass. Configuring diverse distributions and ranges yields 10 datasets with more than 40{,}000 objects.

\noindent\textbf{Baselines.} We compare against two groups. \emph{(i) HOI controllers.} CooHOI~\cite{box_gao2024coohoi} is one of the few controllers that considers weight characteristics in object transportation, so we adopt it as the primary baseline in four configurations: a per-skill carry policy \textbf{Coo(c)} and push--pull policy \textbf{Coo(pp)}, a mass-label-augmented carry policy \textbf{Coo(Aug)}, and the cooperative variant \textbf{Coo(coo)}; we additionally report InterPhys~\cite{InterPhysHassan2023}. \emph{(ii) Adaptation paradigms.} To isolate the contribution of inference-time belief tuning, we compare against three schemes that share our base Hauling Controller but obtain the latent code differently: \textbf{Oracle} encodes the ground-truth context $\mu$ with the trained encoder $\epsilon_c$ (privileged information, no test-time search, a privileged-information reference); \textbf{RMA}~\cite{kumar2021rma} trains a second-stage module that regresses the latent feed-forward from a state--action history; and \textbf{Ours~w/o~AWR} disables the inference-time belief search and uses the prior mean $z\!=\!0$.

\noindent\textbf{Metrics and Implementation.} We report the supported \emph{weight range} (kg), \emph{success rate} (Succ., \%), \emph{precision} (Prec., final ground-point error in cm), and two expressiveness measures. \emph{Average Pairwise Distance} (APD) is a standard motion-diversity score: the mean pairwise Euclidean distance among the root-relative joint positions of the postures collected across successful rollouts (near zero for a deterministic single-posture controller). \emph{MOS} is a user study: $20$ participants rate anonymized, randomly ordered clips on a $1$--$5$ scale for motion naturalness and the plausibility of the displayed effort. Latent tuning uses $K\!=\!3$ trials, each advantage-weighted update from a batch of $512$ parallel-environment rollouts, with learning rate $1\times10^{-3}$.

\subsection{Qualitative Results: Property-Dependent Expressive Exertion}\label{exp:q1}

\begin{figure}[t]
  \centering

  \begin{minipage}{0.58\linewidth}
    \centering
    \includegraphics[width=\linewidth]{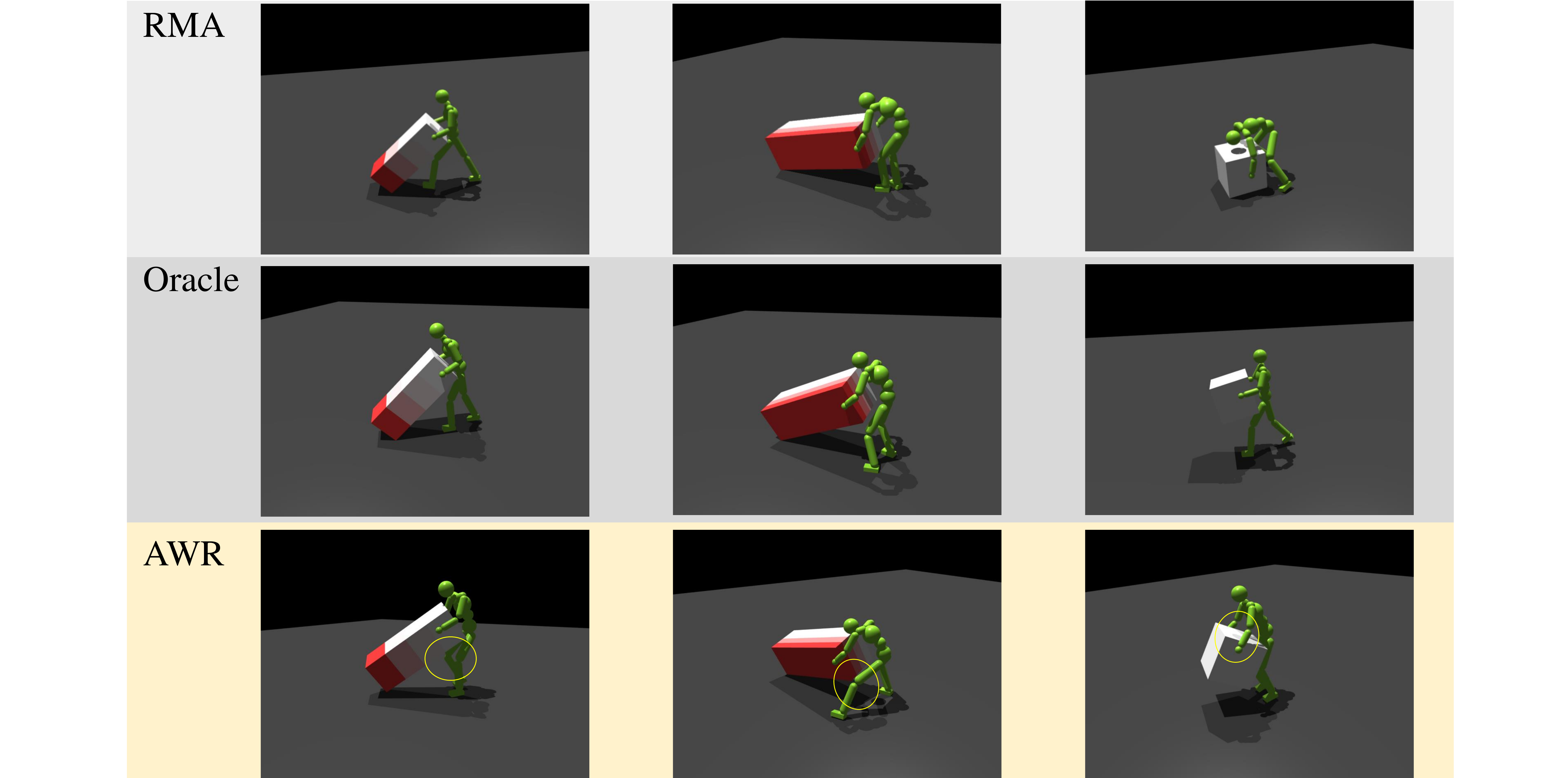}
  \end{minipage}\hfill
  \begin{minipage}{0.42\linewidth}
    \centering
    \includegraphics[width=\linewidth]{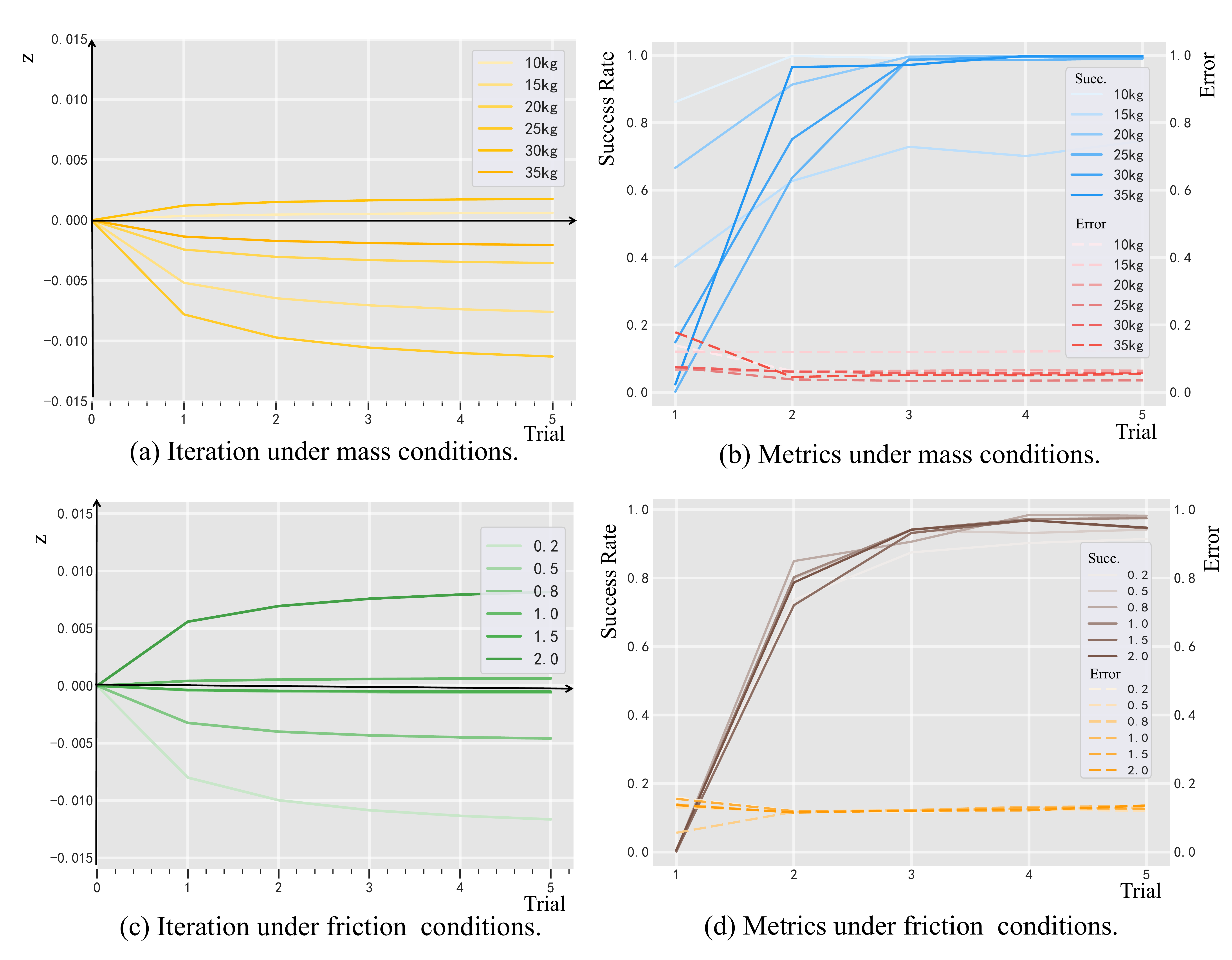}
  \end{minipage}
  \vspace{-10pt}

  \medskip

  \begin{minipage}[t]{0.58\linewidth}
    \captionof{figure}{Exertion postures for the same object.}
    \label{fig:posture_compare}
  \end{minipage}\hfill
  \begin{minipage}[t]{0.42\linewidth}
    \captionof{figure}{AWR iteration and metrics.}
    \label{fig:awr_iteration}
  \end{minipage}
  \vspace{-20pt}
\end{figure}

  \noindent\textbf{Posture comparison across methods.} For the same target object, Fig.~\ref{fig:posture_compare} contrasts the exertion postures of RMA, Oracle, and our AWR-based adaptation. Acting on exact or one-shot-estimated parameters, Oracle
  and RMA settle into near-identical, deterministic postures, or simply fail on the
  first attempt with no chance to correct. Our scheme instead exhibits effortful, expressive motion. The second column makes this visible: facing an object
  whose mass is biased to one side, the character misjudges the center-of-mass offset
  on a given trial, leans into that side, and braces hard with its legs. Light or
  heavy, balanced or eccentric, no single trial is the most accurate estimate of the
  true properties, yet it is this imperfect process that produces visibly distinct,
  physically plausible exertion. In contrast, given exact parameters, Oracle has nothing to adjust, and RMA, with one feed-forward estimate from history, never
  builds such well-calibrated expressiveness over repeated attempts.

\begin{figure}[t]
  \centering
  \includegraphics[width=\linewidth]{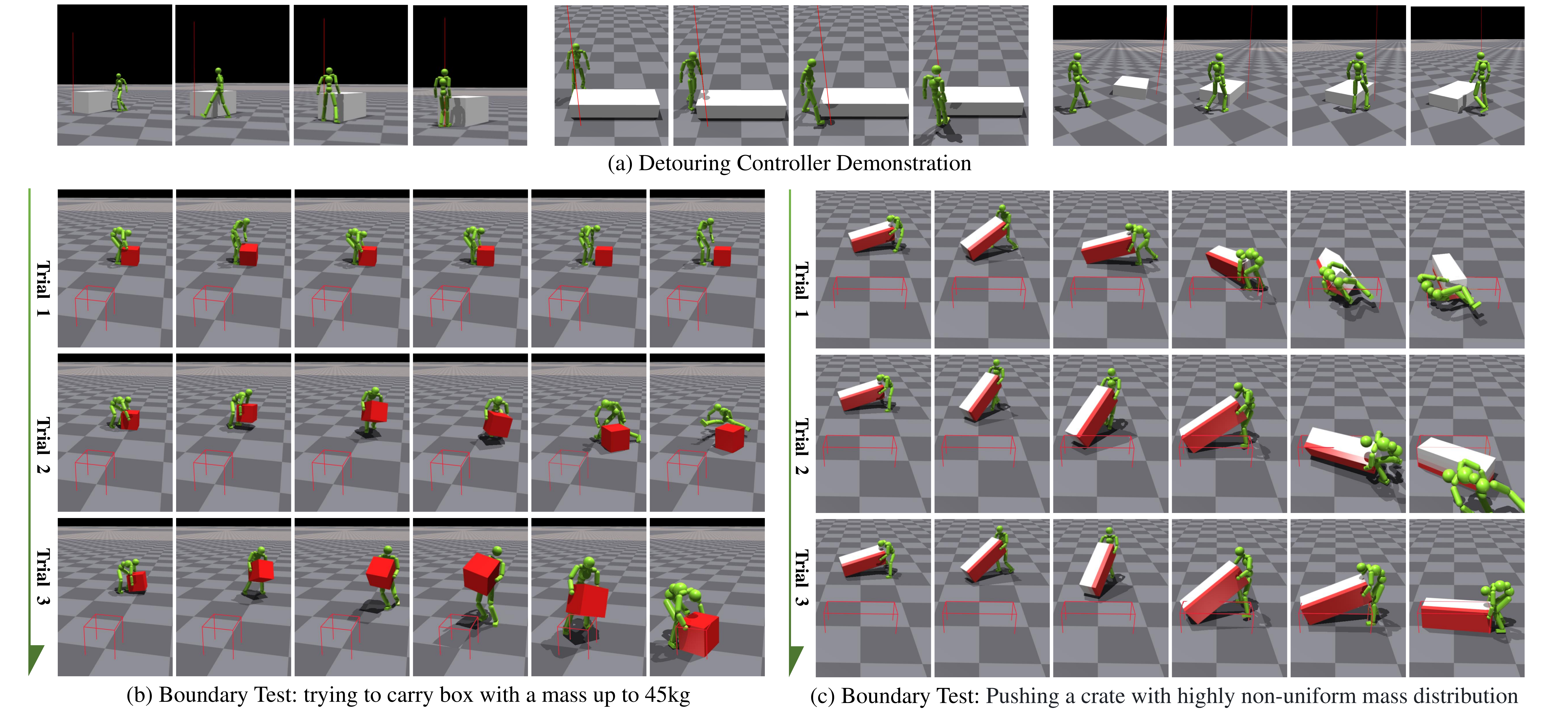}
  \caption{Detouring skill demonstration and adaptation across estimate-tuning iterations}
  \label{fig:trial_iter}
\end{figure}

\noindent\textbf{Controllable adaptation across trials.} Across trials the character acts under a sequence of different, still-imperfect estimations, and the intermediate iterations themselves yield expressive motion. Fig.~\ref{fig:trial_iter} shows the motion of the carry and push--pull skills across different intermediate iterations: the character moves from probing and failing in the early trials to a stable, successful strategy. Each intermediate trial is itself an extractable, expressive exertion clip. For example, in Fig.~\ref{fig:trial_iter}(b) the character attempts to carry a $45$\,kg box; by the third iteration it has successfully adapted to the heavy load, arching its back into a highly expressive interaction posture. Such postures are themselves absent from the reference motions used for training. The same behavior holds near the method's capability boundary: for a strongly eccentric crate ($1.364\times0.45\times0.45$\,m, $41.13$\,kg) whose center of mass is offset $0.14$\,m toward one side and is prone to toppling (Fig.~\ref{fig:trial_iter}(c)), the character likewise progresses from initial failure to a stable strategy that completes the transport, demonstrating that the method remains effective under extreme weight and eccentric mass.

\subsection{Quantitative Results: Effectiveness and Generalization}\label{exp:q2}

\noindent\textbf{Supported load range and precision.} Aligning with real-world human behavior where failures naturally occur, vivid trial-and-error behaviors inevitably incur a mild drop in success rate. Despite this, our method achieves competitive quantitative metrics overall, and it supports an expanded spectrum of object masses while retaining high task success rates and acceptable positioning precision as shown in Tab.~\ref{tab:haulcon}. In the cubic-box transportation task, our method moves a maximum load that is $166.67\%$ higher than CooHOI and $42.86\%$ higher than CooHOI with mass-label augmentation. It achieves this without privileged physical input, relying solely on trial-and-error adaptation. In the extended-crate transportation task, our approach supports the widest weight range.

\begin{table}[t]
    \centering
    \small
    \caption{Supported load range and precision.}
    \label{tab:haulcon}
    \setlength{\tabcolsep}{3pt}
    \begin{tabular}{lcccc}
        \toprule
        \textbf{Method} & \textbf{Cube (kg)} & \textbf{Crate (kg)} & \textbf{Succ.(\%)} & \textbf{Prec.(cm)}\\
        \midrule
        InterPhys~\cite{InterPhysHassan2023} & $[5, 26]$ & --- & 94.30 & 8.30 \\
        Coo(c)~\cite{box_gao2024coohoi}      & $[5, 15]$ & --- & 92.01 & \textbf{3.21} \\
        Coo(Aug)~\cite{box_gao2024coohoi}    & $[4, 28]$ & --- & 91.00 & 3.99 \\
        \textbf{Ours (c)}                    & $\mathbf{[4, 40]}$ & --- & \textbf{98.18} & 5.75 \\
        \midrule
        Coo(pp)~\cite{box_gao2024coohoi}     & --- & $[12, 28]$ & 42.54 & 15.26 \\
        Coo(coo)~\cite{box_gao2024coohoi}    & --- & $[15, 40]$ & 91.22 & \textbf{7.76} \\
        \textbf{Ours (pp)}                   & --- & $\mathbf{[10, 48]}$ & \textbf{94.38} & 9.06 \\
        \bottomrule
    \end{tabular}
    \vspace{-10pt}
\end{table}

    \noindent\textbf{Effectiveness of AWR convergence.} Fig.~\ref{fig:awr_iteration}(a,c) shows the physical latent embeddings mean converging over trials toward clearly different regions for different physical contexts, confirming that the latent space ultimately converges to diverse physical properties; (b,d) report the corresponding success rate and placement error, with the success rate stabilizing above $90\%$ within several trials while maintaining stable interaction.

  \noindent\textbf{Comparison with adaptation paradigms.} Tab.~\ref{tab:compare}
  compares AWR latent adaptation with other ways of recovering the latent context. Oracle, handed the exact physical parameters, reaches the highest success rate. Ours, with the estimation fixed after the trials converge, is essentially on par with it and above RMA, using no privileged input. Success rate alone says little about whether the motion is lively, so we pair it with posture diversity (APD), measured over successful rollouts only so that it reflects the variety of competent motion rather than the spread of failed attempts. Oracle has the lowest APD: with exact parameters it emits one deterministic posture per object, and even its successes look near-identical. RMA is slightly higher, its history-based estimate adding minor fluctuation. Ours~w/o~AWR ($z\!=\!0$) succeeds only $43.5$/$40.5\%$ of the time; without latent tuning the character never adapts to the object, so its variety is unreliable. Ours keeps success high and reaches the highest APD: among motions that succeed, the postures stay
  varied, since the variation comes from the intermediate iterations on the same object. The user study agrees on the full clips: ours scores the best in MOS.

\begin{table}[t]
    \centering
    \caption{Comparison with adaptation paradigms on carry and push--pull.}
    \label{tab:compare}
    \setlength{\tabcolsep}{3pt}
    \begin{tabular}{lcccccc}
        \toprule
        \multirow{2}{*}{\textbf{Method}}
            & \multicolumn{3}{c}{\textbf{Carry}} & \multicolumn{3}{c}{\textbf{Push--Pull}} \\
        \cmidrule(lr){2-4}\cmidrule(lr){5-7}
            & \textbf{Succ. (\%)} & \textbf{APD$\uparrow$} & \textbf{MOS$\uparrow$}
            & \textbf{Succ. (\%)} & \textbf{APD$\uparrow$} & \textbf{MOS$\uparrow$} \\
        \midrule
        Oracle ($\mu$)            & $96.6$ & $0.07$ & $3.3$ & $93.2$ & $0.10$ & $3.2$ \\
        RMA~\cite{kumar2021rma}   & $91.4$ & $0.18$ & $3.0$ & $89.8$ & $0.11$ & $2.9$ \\
        Ours~w/o~AWR ($z\!=\!0$)  & $43.5$ & $0.32$ & $2.1$ & $40.5$ & $0.28$ & $2.0$ \\
        \textbf{Ours}             & $95.9$ & \textbf{0.41} & \textbf{4.2}
                                  & $92.2$ & \textbf{0.38} & \textbf{4.1}\\
        \bottomrule
    \end{tabular}
  \vspace{-5pt}

\end{table}

\noindent\textbf{High-level planner and detouring.} Fig.~\ref{fig:trial_iter}(a) illustrates the smooth motions of DtrCtrl and verifies the effectiveness of this module. We also verify the failure-aware planner on an obstacle-negotiation scenario: a $2.5$\,cm roadblock with $70^{\circ}$ sloped sides, aligned so the object jams when transported directly to the target. As shown in Tab.~\ref{tab:planner}, the standalone InterCtrl, which achieves a $100\%$ success rate without obstacles, drops to $69.53\%$ once the roadblock is introduced; coordinating it with the Detouring Controller (InterCtrl+DtrCtrl) restores success to $81.64\%$. Besides, with the same iteration budget, DtrCtrl raises the detour success rate from $31.31\%$ to $99.51\%$ and cuts the collision rate from $95.09\%$ to $0.03\%$. The ablation shows the effectiveness of the two-step curriculum.

\begin{table}[t]
    \centering
    \caption{High-level planner on obstacle negotiation and the 2-step curriculum ablation.}
    \label{tab:planner}
    \setlength{\tabcolsep}{6pt}
    \begin{tabular}{lcc}
        \toprule
        \textbf{Method} & \textbf{Succ. (\%)} & \textbf{Prec. (cm)} \\
        \midrule
        InterCtrl (no obstacle)   & 100.00 & 3.08 \\
        InterCtrl (with obstacle) & 69.53  & 4.63 \\
        \textbf{InterCtrl+DtrCtrl}    & \textbf{81.64} & 4.91 \\
        \midrule
        \textbf{Variant (DtrCtrl)} & \textbf{Succ. (\%)} & \textbf{Coll. (\%)} \\
        \midrule
        DtrCtrl w/ curriculum      & \textbf{99.51} & \textbf{0.03} \\
        DtrCtrl w/o curriculum     & 31.31 & 95.09 \\
        \bottomrule
    \end{tabular}
  \vspace{-10pt}

\end{table}

\section{Conclusion and Discussion}
We present ExpHOI, which turns the trial-and-error process of adapting to physical uncertainty into expressive animation. Experiments show that ExpHOI generates property-dependent, expressive exertion postures. Its main limitations are that it does not yet model dexterous hand behavior and lacks global navigation or topological planning. Besides, while AWR is a generalizable approach validated on canonical large-object transportation tasks, further evaluation across a broader set of scenarios is still required. All of these directions are promising avenues for future work.

\noindent\textbf{Acknowledgments.} This work was supported in part by the National
Natural Science Foundation of China (Grant No.~62676095) and the Natural Science
Foundation of Jiangsu Province (Grant No.~BK20220127).

\bibliographystyle{splncs04}
\bibliography{main}

@String{CVPR = {CVPR}}

@String{ECCV = {ECCV}}

@String{ICCV = {ICCV}}

@String{TOG = {ACM TOG}}

@article{
	2021-TOG-AMP,
	author = {Peng, Xue Bin and Ma, Ze and Abbeel, Pieter and Levine, Sergey and Kanazawa, Angjoo},
	title = {AMP: Adversarial Motion Priors for Stylized Physics-Based Character Control},
	journal = {ACM Trans. Graph.},
	issue_date = {August 2021},
	volume = {40},
	number = {4},
	month = jul,
	year = {2021},
	articleno = {1},
	numpages = {15},
	doi = {10.1145/3450626.3459670},
	publisher = {ACM},
	address = {New York, NY, USA},
}

@article{basketball_wang2023physhoi,
  author    = {Wang, Yinhuai and Lin, Jing and Zeng, Ailing and Luo, Zhengyi and Zhang, Jian and Zhang, Lei},
  title     = {PhysHOI: Physics-Based Imitation of Dynamic Human-Object Interaction},
  journal   = {arXiv preprint arXiv:2312.04393},
  year      = {2023},
}

@inproceedings{box_gao2024coohoi,
 author = {Gao, Jiawei and Wang, Ziqin and Xiao, Zeqi and Wang, Jingbo and Wang, Tai and Cao, Jinkun and Hu, Xiaolin and Liu, Si and Dai, Jifeng and Pang, Jiangmiao},
 booktitle = {Advances in Neural Information Processing Systems},
 title = {CooHOI: Learning Cooperative Human-Object Interaction with Manipulated Object Dynamics},
 year = {2024}
}

@inproceedings{box_pan2025tokenhsi,
  title={TokenHSI: Unified Synthesis of Physical Human-Scene Interactions through Task Tokenization},
  author={Pan, Liang and Yang, Zeshi and Dou, Zhiyang and Wang, Wenjia and Huang, Buzhen and Dai, Bo and Komura, Taku and Wang, Jingbo},
  booktitle=CVPR,
  year={2025},
}

@inproceedings{christen2022d,
  title={D-grasp: Physically plausible dynamic grasp synthesis for hand-object interactions},
  author={Christen, Sammy and Kocabas, Muhammed and Aksan, Emre and Hwangbo, Jemin and Song, Jie and Hilliges, Otmar},
  booktitle={Proceedings of the IEEE/CVF Conference on Computer Vision and Pattern Recognition},
  pages={20577--20586},
  year={2022}
}

@incollection{dyn_caoHMP2020,
  author = {Zhe Cao and
  Hang Gao and
  Karttikeya Mangalam and
  Qizhi Cai and
  Minh Vo and
  Jitendra Malik},
  title = {Long-term human motion prediction with scene context},
  booktitle = ECCV, 
  year = {2020},
  }

@inproceedings{dyn_li2023controllable,
  title={Controllable human-object interaction synthesis},
  author={Li, Jiaman and Clegg, Alexander and Mottaghi, Roozbeh and Wu, Jiajun and Puig, Xavier and Liu, C. Karen},
  booktitle=ECCV,
  year={2024}
}

@misc{dyn_ron2025hoidinihumanobjectinteractiondiffusion,
      title={HOIDiNi: Human-Object Interaction through Diffusion Noise Optimization}, 
      author={Roey Ron and Guy Tevet and Haim Sawdayee and Amit H. Bermano},
      year={2025},
      eprint={2506.15625},
      archivePrefix={arXiv},
      primaryClass={cs.CV},
}

@INPROCEEDINGS{dyn_TowardsDiverse,
  author={Wang, Jingbo and Rong, Yu and Liu, Jingyuan and Yan, Sijie and Lin, Dahua and Dai, Bo},
  booktitle=CVPR, 
  title={Towards Diverse and Natural Scene-aware 3D Human Motion Synthesis}, 
  year={2022},
  volume={},
  number={},
  pages={20428-20437},
  doi={10.1109/CVPR52688.2022.01981}}

@inproceedings{fatigue_cheema2023discovering,
    title     = {Discovering Fatigued Movements for Virtual Character Animation},
    author    = {Cheema, Noshaba and Xu, Rui and Kim, Nam Hee and H{\"a}m{\"a}l{\"a}inen, Perttu and Golyanik, Vladislav and Habermann, Marc and Theobalt, Christian and Slusallek, Philipp},
    booktitle = {SIGGRAPH Asia 2023 Conference Papers},
    year      = {2023},
    doi       = {10.1145/3610548.3618176}
}

@article{football_Liu2022MotorControl,
  author    = {Liu, Siqi and Lever, Guy and Wang, Zhe and Merel, Josh and Eslami, SM Ali and Hennes, Daniel and Czarnecki, Wojciech M and Tassa, Yuval and Omidshafiei, Shayegan and Abdolmaleki, Abbas and others},
  title     = {From motor control to team play in simulated humanoid football},
  journal   = {Science Robotics},
  year      = {2022},
  volume    = {7},
  number    = {69},
  pages     = {eabo0235},
}

@inproceedings{huang2023compass,
  title={What went wrong? closing the sim-to-real gap via differentiable causal discovery},
  author={Huang, Peide and Zhang, Xilun and Cao, Ziang and Liu, Shiqi and Xu, Mengdi and Ding, Wenhao and Francis, Jonathan and Chen, Bingqing and Zhao, Ding},
  booktitle={Conference on Robot Learning},
  pages={734--760},
  year={2023},
}

@inproceedings{
	InterPhysHassan2023,
	author = {Hassan, Mohamed and Guo, Yunrong and Wang, Tingwu and Black, Michael and Fidler, Sanja and Peng, Xue Bin},
	title = {Synthesizing Physical Character-Scene Interactions},
	year = {2023},
	isbn = {9798400701597},
	publisher = {Association for Computing Machinery},
	address = {New York, NY, USA},
	doi = {10.1145/3588432.3591525},
	booktitle = {ACM SIGGRAPH 2023 Conference Proceedings},
	pages = {1--9},
	articleno = {63},
	numpages = {9},
	location = {Los Angeles, CA, USA},
	series = {SIGGRAPH '23}
}

@article{komura2000musculoskeletal,
    title     = {Creating and retargetting motion by the musculoskeletal human body model},
    author    = {Komura, Taku and Shinagawa, Yoshihisa and Kunii, Tosiyasu L.},
    journal   = {The Visual Computer},
    volume    = {16},
    number    = {5},
    pages     = {254--270},
    year      = {2000},
    publisher = {Springer}
}

@inproceedings{kumar2021rma,
	author = {Kumar, Ashish and Fu, Zipeng and Pathak, Deepak and Malik, Jitendra},
	title = {{RMA}: Rapid Motor Adaptation for Legged Robots},
	booktitle = {Robotics: Science and Systems},
	year = {2021},
	doi = {10.15607/RSS.2021.XVII.011}
}

@inproceedings{
  l_context_xiao2024unified,
  title={Unified Human-Scene Interaction via Prompted Chain-of-Contacts},
  author={Zeqi Xiao and Tai Wang and Jingbo Wang and Jinkun Cao and Wenwei Zhang and Bo Dai and Dahua Lin and Jiangmiao Pang},
  booktitle={The Twelfth International Conference on Learning Representations},
  year={2024}
}

@inproceedings{l_pan2024synthesizing,
  title={Synthesizing physically plausible human motions in 3d scenes},
  author={Pan, Liang and Wang, Jingbo and Huang, Buzhen and Zhang, Junyu and Wang, Haofan and Tang, Xu and Wang, Yangang},
  booktitle={2024 International Conference on 3D Vision (3DV)},
  pages={1498--1507},
  year={2024},
  organization={IEEE}
}

@inproceedings{l_wang2025sims,
title = {SIMS: Simulating Stylized Human-Scene Interactions with Retrieval-Augmented Script Generation},
booktitle = ICCV,
author = {Wenjia Wang and Liang Pan and Zhiyang Dou and Jidong Mei and Zhouyingcheng Liao and Yuke Lou and Yifan Wu and Lei Yang and Jingbo Wang and Taku Komura},
year = {2025}
}

@article{MetaStrategy,
title	= {Learning Fast Adaptation with Meta Strategy Optimization},
author	= {Wenhao Yu and Jie Tan and Yunfei Bai and Erwin Johan Coumans and Sehoon Ha},
journal = {IEEE Robotics and Automation Letters},
volume  = {5},
number  = {2},
pages   = {2950--2957},
year	= {2020},
publisher = {IEEE}}

@article{NavRL,
  author={Xu, Zhefan and Han, Xinming and Shen, Haoyu and Jin, Hanyu and Shimada, Kenji},
  journal={IEEE Robotics and Automation Letters}, 
  title={NavRL: Learning Safe Flight in Dynamic Environments}, 
  year={2025},
  volume={10},
  number={4},
  pages={3668-3675},
  doi={10.1109/LRA.2025.3546069}}

@article{peng2019awr,
	author = {Peng, Xue Bin and Kumar, Aviral and Zhang, Grace and Levine, Sergey},
	title = {Advantage-Weighted Regression: Simple and Scalable Off-Policy Reinforcement Learning},
	journal = {arXiv preprint arXiv:1910.00177},
	year = {2019}
}

@article{peng2022ase,
  title={Ase: Large-scale reusable adversarial skill embeddings for physically simulated characters},
  author={Peng, Xue Bin and Guo, Yunrong and Halper, Lina and Levine, Sergey and Fidler, Sanja},
  journal={ACM Transactions On Graphics (TOG)},
  volume={41},
  number={4},
  pages={1-17},
  year={2022},
  publisher={ACM New York, NY, USA}
}

@InProceedings{pmlr-v70-finn17a,
  title = 	 {Model-Agnostic Meta-Learning for Fast Adaptation of Deep Networks},
  author =       {Chelsea Finn and Pieter Abbeel and Sergey Levine},
  booktitle = 	 {Proceedings of the 34th International Conference on Machine Learning},
  pages = 	 {1126--1135},
  year = 	 {2017},
  editor = 	 {Precup, Doina and Teh, Yee Whye},
  volume = 	 {70},
  series = 	 {Proceedings of Machine Learning Research},
  month = 	 {06--11 Aug},
  publisher =    {PMLR},
}

@inproceedings{ramos2019bayessim,
  author    = {Fabio Ramos and Rafael Possas and Dieter Fox},
  title     = {BayesSim: Adaptive Domain Randomization Via Probabilistic Inference for Robotics Simulators},
  booktitle = {Robotics: Science and Systems (RSS)},
  year      = {2019},
  doi       = {10.15607/RSS.2019.XV.029}
}

@inproceedings{
	RoboImitationPeng20,
	author = {Peng, Xue Bin and Coumans, Erwin and Zhang, Tingnan and Lee, Tsang-Wei Edward and Tan, Jie and Levine, Sergey},
	booktitle={Robotics: Science and Systems},
	year = {2020},
	month = {07},
	title = {Learning Agile Robotic Locomotion Skills by Imitating Animals},
	doi = {10.15607/RSS.2020.XVI.064}
}

@misc{schulman2017proximal,
  title={Proximal policy optimization algorithms},
  author={Schulman, John and Wolski, Filip and Dhariwal, Prafulla and Radford, Alec and Klimov, Oleg},
  archivePrefix = {arxiv},
  eprint={1707.06347},
  year={2017}
}

@inproceedings{sim_NEURIPSDATASETSANDBENCHMARKS2021_28dd2c79,
 author = {Makoviychuk, Viktor and Wawrzyniak, Lukasz and Guo, Yunrong and Lu, Michelle and Storey, Kier and Macklin, Miles and Hoeller, David and Rudin, Nikita and Allshire, Arthur and Handa, Ankur and State, Gavriel},
 booktitle = {Proceedings of the Neural Information Processing Systems Track on Datasets and Benchmarks},
 editor = {J. Vanschoren and S. Yeung},
 pages = {},
 title = {Isaac Gym: High Performance GPU Based Physics Simulation For Robot Learning},
 volume = {1},
 year = {2021}
}

@article{starke2019neural,
  title={Neural state machine for character-scene interactions},
  author={Starke, Sebastian and Zhang, He and Komura, Taku and Saito, Jun},
  journal={ACM Transactions on Graphics},
  volume={38},
  number={6},
  pages={178},
  year={2019},
}

@article{
  tennis_zhang2023vid2player3d,
  author = {Zhang, Haotian and Yuan, Ye and Makoviychuk, Viktor and Guo, Yunrong and Fidler, Sanja and Peng, Xue Bin and Fatahalian, Kayvon},
  title = {Learning Physically Simulated Tennis Skills from Broadcast Videos},
  journal = {ACM Trans. Graph.},
  issue_date = {August 2023},
  numpages = {14},
  doi = {10.1145/3592408},
  publisher = {ACM},
  address = {New York, NY, USA},
  year = {2023},
}

@inproceedings{tessler2023calm,
  title={Calm: Conditional adversarial latent models for directable virtual characters},
  author={Tessler, Chen and Kasten, Yoni and Guo, Yunrong and Mannor, Shie and Chechik, Gal and Peng, Xue Bin},
  booktitle={ACM SIGGRAPH 2023 Conference Proceedings},
  pages={1-9},
  year={2023}
}

@misc{ttennis_su2025hitterhumanoidtabletennis,
      title={HITTER: A HumanoId Table TEnnis Robot via Hierarchical Planning and Learning}, 
      author={Zhi Su and Bike Zhang and Nima Rahmanian and Yuman Gao and Qiayuan Liao and Caitlin Regan and Koushil Sreenath and S. Shankar Sastry},
      year={2025},
      eprint={2508.21043},
      archivePrefix={arXiv},
      primaryClass={cs.RO},
}

@inproceedings{xu2023interdiff,
  title={Interdiff: Generating 3d human-object interactions with physics-informed diffusion},
  author={Xu, Sirui and Li, Zhengyuan and Wang, Yu-Xiong and Gui, Liang-Yan},
  booktitle=ICCV,
  pages={14928--14940},
  year={2023}
}

@inproceedings{xu2025interact,
  title = {{InterAct}: Advancing Large-Scale Versatile 3D Human-Object Interaction Generation},
  author = {Xu, Sirui and Li, Dongting and Zhang, Yucheng and Xu, Xiyan and Long, Qi and Wang, Ziyin and Lu, Yunzhi and Dong, Shuchang and Jiang, Hezi and Gupta, Akshat and Wang, Yu-Xiong and Gui, Liang-Yan},
  booktitle = CVPR,
  year = {2025},
}

@inproceedings{xu2025intermimic,
  title = {{InterMimic}: Towards Universal Whole-Body Control for Physics-Based Human-Object Interactions},
  author = {Xu, Sirui and Ling, Hung Yu and Wang, Yu-Xiong and Gui, Liang-Yan},
  booktitle = CVPR,
  year = {2025},
}

@misc{zhang2024force,
  title={Force: Dataset and method for intuitive physics guided human-object interaction},
  author={Zhang, Xiaohan and Bhatnagar, Bharat Lal and Starke, Sebastian and Petrov, Ilya and Guzov, Vladimir and Dhamo, Helisa and P{\'e}rez-Pellitero, Eduardo and Pons-Moll, Gerard},
  year={2024},
  eprint       = {2403.11237},
  archivePrefix = {arXiv},
  primaryClass = {cs.CV}
}
\end{document}